\documentclass[sigconf, authorversion]{acmart}
\usepackage[utf8]{inputenc}
\usepackage{xspace}
\usepackage{amsmath}
\usepackage{tikz}
\usepackage{pgfplots}
\usepackage{standalone}
\usepackage{tabularx}
\usepackage{array}
\usepackage{booktabs}
\usepackage[justification=centering, caption=false]{subfig}
\usepackage{algorithm}
\usepackage[noend]{algpseudocode}
\usepackage{enumitem}
\usepackage{calc}
\usepackage{threeparttable}
\usepackage{url}
\usepackage{hyperref}
\newcolumntype{C}{>{\centering\arraybackslash}X}

\usetikzlibrary{calc}
\usetikzlibrary{fit}
\usetikzlibrary{shapes}
\usetikzlibrary{positioning}
\usetikzlibrary{matrix}

\newlength\eltWidth
\newlength\eltHeight
\def\horSep{0.5em}
\def\vertSep{\horSep}
\def\labelSep{2*\horSep}

\tikzstyle{elt} = [
  inner sep=0pt,
  outer sep=0pt,
  text width=\eltWidth,
  font=\small,
  align=center
]

\tikzstyle{register} = [
  rectangle split,
  minimum height=\eltHeight,
  rectangle split horizontal=true,
  rectangle split parts=5,
  rectangle split part align=base,
  draw
]

\tikzstyle{operator} = [
  chamfered rectangle,
  font=\small,
  text width=13em,
  text height=1.2ex,
  text depth=0.1ex,
  align=center,
  draw
]

\tikzstyle{label} = [
    font=\small,
    inner sep=0,
    outer sep=0
]

\newlength\cellHeight
\newlength\cellWidth
\tikzset{
  memory cell/.style={
    rectangle,
    draw,
    font=\small,
    text depth=0.25ex,
    text height=\cellHeight,
    text width=\cellWidth,
    align=center,
  },      
  memory/.style={
    matrix of math nodes,
    row sep=-\pgflinewidth,
    column sep=-\pgflinewidth,
    nodes={
        memory cell
    },
    execute at empty cell={\node[draw=none]{};}
  },
  memory multicell/.style={
    draw,
    inner ysep=\pgflinewidth/2,
    inner xsep=-\pgflinewidth/2,
    font=\small,
    align=center,
    text height=\cellHeight,
    text depth=0.1ex,
  }
}

\tikzstyle{memory label} = [
    font=\small
]

\pgfplotsset{compat=1.12}
\pgfplotsset{small fonts/.style={
  title style = {font=\small, yshift=-0.5em},
  label style = {font=\small},
  tick label style = {font=\small},
  legend style={font=\small}
}}
\pgfplotsset{recall plot/.style={
  small fonts,
  xtick={1,2,5,10,20,50,100,200,500,1000},
  xticklabels={1,2,5,10,20,50,100,200,500,1K},
  xlabel=$R$,
  ylabel=Recall@$R$,
  legend pos=south east,
  legend cell align=left,
}}
\pgfplotsset{time plot/.style={
  small fonts,
  yticklabel=\empty,
  legend pos=south east,
  legend cell align=left,
  xlabel={Total query time [ms]},
  every node near coord/.append style={font=\small},
  nodes near coords align={horizontal},
  every node near coord/.append style={xshift=0.05em},
  nodes near coords={
        \pgfmathprintnumber[fixed relative, precision=2]{\pgfplotspointmeta}
  }
}}
\pgfplotsset{time bar/.style={
  xbar, point meta=explicit,
  error bars/.cd,
  x dir=both,
  x explicit
}}

\newcommand*{\eg}{e.g.,\xspace}
\newcommand*{\ie}{i.e.,\xspace}

\newcommand*{\stimes}{{\times}}

\newcommand*{\q}[1][]{%
  \ifthenelse{\isempty{#1}}%
  {\operatorname{q}}%
  {\operatorname{q_#1}}%
}

\newcommand*{\uv}[1][]{%
  \ifthenelse{\isempty{#1}}%
  {\operatorname{u}}%
  {\operatorname{u_#1}}%
}

\DeclareMathOperator*{\argmin}{arg\,min}

\begin{document}

\title{Accelerated Nearest Neighbor Search with Quick ADC}

\author{Fabien André}
\affiliation{
  \institution{Technicolor}
}
\orcid{0000-0001-8620-7632}
\email{fabien.andre@technicolor.com}

\author{Anne-Marie Kermarrec}
\affiliation{
  \institution{Inria}
}
\email{anne-marie.kermarrec@inria.fr}

\author{Nicolas Le Scouarnec}
\affiliation{
  \institution{Technicolor}
}
\email{nicolas.lescouarnec@technicolor.com}

\copyrightyear{2017}
\acmYear{2017}
\setcopyright{licensedothergov}
\acmConference{ICMR '17}{June 06-09, 2017}{Bucharest, Romania}
\acmPrice{15.00}
\acmDOI{http://dx.doi.org/10.1145/3078971.3078992}
\acmISBN{978-1-4503-4701-3/17/06}

\begin{abstract}
Efficient Nearest Neighbor (NN) search in high-dimensional spaces is a foundation of many multimedia retrieval systems. Because it offers low responses times, Product Quantization (PQ) is a popular solution. PQ compresses high-dimensional vectors into short codes using several sub-quantizers, which enables in-RAM storage of large databases. This allows fast answers to NN queries, without accessing the SSD or HDD. The key feature of PQ is that it can compute distances between short codes and high-dimensional vectors using cache-resident lookup tables. The efficiency of this technique, named Asymmetric Distance Computation (ADC), remains limited because it performs many cache accesses.

In this paper, we introduce Quick ADC, a novel technique that achieves a 3 to 6 times speedup over ADC by exploiting Single Instruction Multiple Data (SIMD) units available in current CPUs. Efficiently exploiting SIMD requires algorithmic changes to the ADC procedure. Namely, Quick ADC relies on two key modifications of ADC: (i) the use 4-bit sub-quantizers instead of the standard 8-bit sub-quantizers and (ii) the quantization of floating-point distances. This allows Quick ADC to exceed the performance of state-of-the-art systems, e.g., it achieves a Recall@100 of 0.94 in 3.4 ms on 1 billion SIFT descriptors (128-bit codes).

\end{abstract}


\maketitle

\section{Introduction}
The Nearest Neighbor (NN) search problem consists in finding the closest vector $x$ to a query vector $y$ among a database of $N$ $d$-dimensional vectors. Efficient NN search in high-dimensional spaces is a requirement in many multimedia retrieval applications, such as image similarity search, image classification, or object recognition. These problems typically involve extracting high-dimensional feature vectors, or descriptors, and finding the NN of the extracted descriptors among a database of descriptors. For images, SIFT \cite{Lowe1999} and GIST descriptors~\cite{Oliva2001} are commonly used.

Although efficient NN search solutions have been proposed for low-dimensional spaces, \emph{exact} NN search remains challenging in high-dimensional spaces due to the notorious curse of dimensionality. As a consequence, much research work has been devoted to Approximate Nearest Neighbor (ANN) search. ANN search returns sufficiently close neighbors instead of the exact NN. Product Quantization (PQ) \cite{Jegou2011} is a widely used~\cite{Krapac2014,Xie2015} ANN search approach. PQ compresses high-dimensional vectors into short codes of a few bytes, enabling in-RAM storage of large databases. This allows fast answers to ANN queries, without SSD or HDD accesses.

The key feature of PQ is that it allows computing distances between uncompressed query vectors and compressed database vectors. This technique, known as Asymmetric Distance Computation (ADC), relies on cache-resident lookup tables. Although ADC is faster than distance computations in high-dimensional spaces, its efficiency remains low because it performs many cache accesses. To date, much of the research work has been devoted to the development of efficient inverted indexes~\cite{Babenko2015, Xia2013}, which reduce the number of ADCs required to answer NN queries. Recently, there also has been an interest in increasing the performance of the ADC procedure itself with the introduction of PQ Fast Scan~\cite{Andre2015}. Unfortunately, PQ Fast Scan cannot be combined with efficient inverted indexes, limiting its usefulness in practical cases. In this paper, we introduce Quick ADC, a high-performance ADC procedure that can be combined with inverted indexes. More specifically, this paper makes two contributions, detailed in the next two paragraphs.

First, we detail the design of Quick ADC. Like PQ Fast Scan, Quick ADC replaces cache accesses by SIMD in-register shuffles to accelerate the ADC procedure. Exploiting SIMD in-register shuffles requires storing the lookup tables used by the ADC procedure in SIMD registers. However, these registers are much smaller than the lookup tables used by the conventional ADC procedure. Therefore, algorithmic changes are required to obtain small lookup tables that fit SIMD registers. PQ Fast Scan obtains such small lookup tables by grouping the codes of the database. This approach prevents PQ Fast Scan from being combined with inverted indexes. Quick ADC takes a different approach to obtain small lookup tables, which is compatible with inverted indexes. Namely, Quick ADC relies on two key ideas: (i) the use of 4-bit sub-quantizers, instead of the standard 8-bit sub-quantizers, and (ii) the quantization of floating-point distances to 8-bit integers.

Second, we implement Quick ADC and evaluate its performance in a wide range of scenarios.
It is known that the use of 4-bit quantizers instead of the common 8-bit quantizer can cause a loss of recall~\cite{Jegou2011}. However, we show that this loss is small or negligible, especially when combining Quick ADC with inverted indexes and Optimized Product Quantization (OPQ), a variant of PQ. On the SIFT1B dataset, Quick ADC achieves a better speed-accuracy tradeoff than the state-of-art OMulti-D-OADC system~\cite{Ge2014, Babenko2015}, e.g., Quick ADC achieves a Recall@100 of 0.94 in 3.4 ms (128-bit codes).

\section{Background}
In this section, we describe how Product Quantizers (PQ) and Optimized Product Quantizers (OPQ) encode vectors into short codes. We then detail the ANN search process in databases of short codes. Lastly, we analyze the impact of PQ parameters on ANN search speed and recall.

\subsection{Vector Encoding}
\label{sec:vecenc}
\emph{Vector Quantizers.} To encode vectors as short codes, PQ builds on vector quantizers.  A vector quantizer, or quantizer, is a function $\operatorname{q}$ which maps a vector $x \in \mathbb{R}^d$, to a vector $c_i \in \mathbb{R}^d$ belonging to a predefined set of vectors $\mathcal{C}$. Vectors $c_i$ are called \emph{centroids}, and the set of centroids $\mathcal{C}$, of cardinality $k$, is the \emph{codebook}. For 
a given codebook $\mathcal{C}$, a quantizer which minimizes the quantization error must satisfy Lloyd's condition and map the vector $x$ to its closest centroid $c_i$:
\[
\operatorname{q}(x) = \argmin_{c_i \in \mathcal{C}}{||x-c_i||}.
\]
A vector quantizer can be used to encode a vector $x \in \mathbb{R}^d$ into a short code $i \in \{0 \dots k-1\}$ using the encoder $\operatorname{enc}$:
\[
\operatorname{enc}(x) = i,\text{ such that }\operatorname{q}(x) = c_i
\]
The short code $i$ only occupies $b = \lceil \log_2(k) \rceil$ bits, which is typically much lower the $d\cdot32$ bits occupied by a vector $x \in \mathbb{R}^d$ stored as an array of $d$ single-precision floats (32 bit each). To maintain the quantization error low enough for ANN search, a very large codebook \eg $k=2^{64}$ or $k=2^{128}$ is required. However, training such codebooks is not tractable both in terms of processing and memory requirements.

\emph{Product Quantizers.} Product quantizers overcome this issue by dividing a vector $x \in \mathbb{R}^d$ into $m$ sub-vectors, $x = (x^0,\dots,x^{m-1})$, assuming that $d$ is a multiple of $m$. Each sub-vector $x^j \in \mathbb{R}^{d/m}$, $j \in \{0,\dots,m-1\}$ is quantized using a sub-quantizer $\operatorname{q}^j$. Each sub-quantizer $\operatorname{q}^j$ has a distinct codebook $\mathcal{C}^j=(c_i^j)_{i=0}^{k-1}$ of cardinality $k$. A product quantizer $\operatorname{pq}$ maps a vector $x \in \mathbb{R}^d$ as follows:
\begin{align*}
\operatorname{pq}(x) &= \left ( \operatorname{q}^0(x^0),\dots,\operatorname{q}^{m-1}(x^{m-1}) \right )\\
&= (c^0_{i_0},\dots,c^{m-1}_{i_{m-1}})
\end{align*}
The codebook $\mathcal{C}$ of the product quantizer $q$ is given by the cartesian product of the sub-quantizers codebooks:
\[
\mathcal{C} = \mathcal{C}^0 \times \dots \times \mathcal{C}^{m-1}
\]
The cardinality of the product quantizer codebook $\mathcal{C}$ is $k^m$. Thus, a product quantizer is able to produce a large number of centroids $k^m$
while only requiring storing and training $m$ codebooks of cardinality $k$. A product quantizer can be used to encode a vector $x$ into a short code, by concatenating codes produced by sub-quantizers:
\[
\operatorname{enc}(x) = (i_0,\dots,i_{m-1}),\text{ such that }\operatorname{q}(x) = (c^0_{i_0},\dots,c^{m-1}_{i_{m-1}})
\]
The short code $(i_0,\dots,i_{m-1})$ requires $ \lceil \log_2(k^m) \rceil = m \cdot b$ bits of storage, where $b = \lceil \log_2(k) \rceil $.

\emph{Optimized Product Quantizers.} Cartesian k-means (CKM) \cite{Norouzi2013} and Optimized Product Quantizers (OPQ) \cite{Ge2014} and  optimize the sub-space decomposition by multiplying the vector $x$ by an orthonormal matrix $R \in \mathbb{R}^{d \times d}$ before quantization. The matrix $R$ allows for arbitrary rotation and permutation of vector components. An optimized product quantizer $\operatorname{opq}$ maps a vector $x$ as follows:
\[
\operatorname{opq}(x) = \operatorname{pq}(Rx), \text{ such that } R^TR=I,
\]
where $\operatorname{pq}$ is a product quantizer. Optimized product quantizers can be used to encode vectors into short codes like product quantizers.

\subsection{Inverted Indexes}
\label{sec:ivf}
The simplest search strategy, exhaustive search, involves encoding database vectors as short codes using PQ or OPQ and storing short codes in RAM. At query time, the whole database is scanned for nearest neighbors.

The more refined non-exhaustive search strategy relies on inverted indexes (or IVF)~\cite{Jegou2011, Jegou2011R} to avoid scanning the whole database. An inverted index uses a quantizer $\operatorname{q_i}$ to partition the input vector space into $K$ Voronoi cells. Vectors lying in each cell are stored in an inverted list. At query time, the inverted index is used to find the closest cells to the query vector, which are then scanned. Inverted indexes therefore offer a lower query response time. When adding a vector $x$ to an indexed database, its residual $\operatorname{r}(x)$ is first computed:
\[
\operatorname{r}(x) = x - \operatorname{q_i}(x)
\]

The residual $\operatorname{r}(x)$ is then encoded into a short code using a product quantizer. This code is then stored in the appropriate inverted list of the inverted index. Indexed databases therefore use two quantizers: a quantizer for the index ($\operatorname{q_i}$) and a product quantizer to encode residuals into short codes. The energy of residuals $\operatorname{r}(x)$ is smaller than the energy of input vectors $x$, thus there is a lower quantization error when encoding residuals into short codes. Non-exhaustive search therefore offers a higher recall than exhaustive search in addition to the lower response time. Inverted indexes however incur a memory overhead (usually 4 bytes per database vector). This memory overhead is negligible in the case of small databases ($\sim4$MB for 1 million vectors) and for large databases, exhaustive search is anyway hardly tractable. Non-exhaustive search is therefore preferred to exhaustive search in most cases.

\subsection{ANN Search}

\begin{algorithm}
  \caption{ANN Search}\label{alg:ann}
  \begin{algorithmic}[1]
    \Function{nns}{$\{\mathcal{C}^j\}_{j=0}^{m}, \mathit{database}, y, R$}
    \State $\mathit{list}, y' \gets $\Call{index\_get\_list}{$\mathit{database}, y$}
    \State $\{D^j\}_{j=0}^{m} \gets$ \Call{compute\_tables}{$y', \{\mathcal{C}^j\}_{j=0}^{m}$} 
    \State \Return{\Call{scan}{$\mathit{list}, \{D^j\}_{j=0}^{m}$}}
    \EndFunction

    \Function{scan}{$\mathit{list}, \{D^j\}_{j=0}^{m}, R$}
    \State $\mathit{neighbors} \gets \operatorname{binheap}(R)$\Comment{binary heap of size $R$} \label{alg:line:binheap}
    \For{$i \gets 0$ to $|\mathit{list}|-1$}  \label{alg:line:iter}
    \State $c \gets \mathit{list}[i]$\Comment{$i$th pqcode}
    \State $d \gets$ \Call{adc}{$p, \{D^j\}_{j=0}^{m}$} \label{alg:line:dist}
    \State $\mathit{neighbors}.\operatorname{add}((i, d))$
    \EndFor
    \State \Return{$\mathit{neighbors}$}
    \EndFunction

    \Function{adc}{$c, \{D^j\}_{j=0}^{m-1}$} \label{alg:line:adc}
    \State $d \gets 0$
    \For{$j \gets 0$ to $m$}
    \State $d \gets d + D^j[c[j]]$ \label{alg:line:ops}
    \EndFor
    \Return{$d$}
    \EndFunction
  \end{algorithmic}
\end{algorithm}

\label{sec:annsearch}
ANN search in a database of short codes consists in three steps: \emph{Index}, which involves retrieving inverted lists from the index, \emph{Tables}, which involves computing lookup tables to speed up distance computations and \emph{Scan} which involves computing distances between the query vector and short codes using the pre-computed lookup tables. Obviously, the step \emph{Index} is only required for non-exhaustive search, and is skipped in the case of exhaustive search. We detail these three steps in the three following paragraphs.

\emph{Index.} In this step, the Voronoi cell of the inverted index quantizer $\operatorname{q_i}$ in which the query vector $y$ lies is determined. The residual $\operatorname{r}(y)$ of the query vector is also computed. In practice, to improve recall, the $\mathit{ma}$ closest cells (typically, $\mathit{ma}=8$ to $64$) are selected. For the sake of simplicity, this section describes the ANN search process for $\mathit{ma}=1$, but each operation is repeated $\mathit{ma}$ times: $\mathit{ma}$ cells are selected, $\mathit{ma}$ sets of lookup tables are computed and $\mathit{ma}$ cells are searched. In the case of exhaustive search no residual is computed and the query vector is used as-is. In the remainder of this section, $y'=\operatorname{r}(y)$ for non-exhaustive search, and $y'=y$ for exhaustive search.

\emph{Tables.} In this step, a set of $m$ lookup tables are computed $\{D^j\}_{j=0}^{m}$, where $m$ is the number of sub-quantizers of the product quantizer. The $j$th lookup table comprises the distance between the $j$ sub-vector of $y'$ and all centroids of the $j$th sub-quantizer:
\begin{equation}
D^j = \left( \left \lVert {y'}^j - \mathcal{C}^j[0] \right\rVert^2,\dots, \left \lVert {y'}^j - \mathcal{C}^j[k-1] \right \rVert^2\right) \label{eqn:tables}
\end{equation}
\emph{Scan.} In this step, the cells of the inverted index selected during the step \emph{Index} are searched for nearest neighbors. This requires computing the distance between the query vectors and short codes using Asymmetric Distance Computation (ADC). ADC computes the distance between the query vector $y$ and a short code $c$ as follows:
\begin{equation}
\operatorname{adc}(y,c) = \sum_{j=0}^{m-1} D^{j}[c[j]] \label{eqn:adc1}
\end{equation}
Equation \ref{eqn:adc1} is equivalent to:
\begin{equation}
\operatorname{adc}(y,c) = \sum_{j=0}^{m-1} \left \lVert {y'}^j - \mathcal{C}^j[c[j]] \right \rVert^2
\end{equation}
Thus, ADC computes the distance between a query vector $y'$ and a code $c$ by summing the distances between the sub-vectors of $y'$ and centroids associated with code $c$ in the $m$ sub-spaces of the product quantizer. When the number of codes in cells is large compared to $k$, the number of centroids of sub-quantizers, using lookup tables avoids computing $\lVert {y'}^j - \mathcal{C}^j[i] \rVert^2$ for the same $i$ multiple times. Thus, lookup tables therefore provide a significant speedup. While scanning inverted lists, neighbors and their associated distances are stored in a binary heap of size $R$ (Algorithm~\ref{alg:ann}, line~\ref{alg:line:binheap}).

\subsection{Impact of PQ Parameters}
\label{sec:paramimpact}

\begin{table}
  \label{tbl:confspeed}
  \caption{Speed-Accuracy tradeoff for 64-bit codes (SIFT1M, Exhaustive search)\label{tbl:tblsize}}
  \begin{tabularx}{\linewidth}{llllXX}
    \toprule
    $m\stimes b$ & Size & Cache & R@100 & Tables & Scan \\
    \midrule
    $16\stimes 4$ & 1 KiB & L1 & 83.1\% & 0.001 ms & 6.1 ms\\
    $8\stimes 8$ & 8 KiB  & L1 & 91.6\% & 0.005 ms & 2.7 ms\\
    $4\stimes 16$ & 1 MiB & L3 & 96.5\% & 0.77 ms& 7.8 ms\\
    \bottomrule
  \end{tabularx}
\end{table}

The two parameters of a product quantizer, $m$, the number of sub-quantizers and $k$, the number of centroids of each sub-quantizer impact: (1) the memory usage of codes, (2) the recall of ANN search and (3) search speed.
In practice, 64-bit codes ($2^{64}$ centroids) or 128-bit codes ($2^{128}$ centroids) are used in most cases.

The second tradeoff is between ANN accuracy and search speed. For a constant memory budget of $m\cdot b$ bits per code, the respective values of $m$ and $b$ impact accuracy and speed. Decreasing $m$, which implies increasing $b$, increases accuracy~\cite{Jegou2011}. We discuss the effect of $m$ and $b$ on the time cost of the \emph{Tables} and \emph{Scan} steps of ANN search (Section~\ref{sec:annsearch}). Each lookup table requires $k=2^{b}$ $l_2$-norm computations in sub-spaces of dimensionality $d/m$. Thus, the complexity of computing all $m$ lookup tables is $O(m\cdot2^{b}\cdot d/m) = O(2^{b}\cdot d)$, and increases exponentially with $b$. In conclusion, decreasing $m$ makes the \emph{Tables} step more costly.

During the \emph{Scan} step, each Asymmetric Distance Computation (ADC) (Algorithm \ref{alg:ann}, line \ref{alg:line:adc}) requires $m$ accesses to lookup tables and $m$ additions (Algorithm \ref{alg:ann}, line \ref{alg:line:ops}). Therefore, decreasing $m$ decreases the number of operations required for each ADC, which is beneficial for search speed. However, decreasing $m$ implies increasing $b$, and thus increasing the size of lookup tables. The size of all lookup tables $\{D^j\}_{j=0}^{m}$ is $m\cdot k \cdot \operatorname{sizeof}(\mathrm{float}) = m \cdot 2^{b} \cdot 4$. It increases linearly with $m$ and exponentially with $b$. Thus, decreasing $m$ increases the size of lookup tables. As the size of lookup tables increases, they need to be stored in larger and slower cache levels which is detrimental to performance~\cite{Andre2015}. In conclusion, decreasing $m$, makes the \emph{Tables} step less costly, except if it causes lookup tables to be stored in slower cache.

To illustrate this, we measure the recall (R@100) and the time cost of the \emph{Tables} and \emph{Scan} steps of ANN search for different $m{\stimes}b$ configurations producing 64-bit codes (Table~\ref{tbl:tblsize}). For $16{\stimes}4$ and $8{\stimes}8$, tables fit the L1 cache. The $8{\stimes}8$ configuration has a lower \emph{Scan} time because it requires less additions and less accesses to lookup tables. The $4{\stimes}16$ configuration requires even less additions and table accesses but lookup tables are stored in the much slower L3 cache. Overall, the $4{\stimes}16$ configuration therefore has a higher \emph{Scan} time. In all cases, the time cost of the \emph{Tables} step increases with $b$.

\section{Quick ADC}
\subsection{Overview}

The performance gains of Quick ADC are achieved by exploiting SIMD. Single Instruction Multiple Data (SIMD) instructions perform the same operation \eg additions, on multiple data elements in one instruction. Consequently, SIMD enables large performance improvements. Thus, optimized linear algebra libraries rely on SIMD to offer high performance. Current CPUs include an SIMD unit in each core. SIMD therefore offers an additional level of parallelism over multi-core processing. ANN search parallelizes naturally over multiple cores by processing a distinct query on each core. With Quick ADC, we propose further increasing performance by speeding up ADC for each query, thanks to the use of SIMD.
To process multiple data elements at once, SIMD instructions operate on wide registers. SSE instructions use 128-bit registers, while the newer AVX instructions use 256-bit registers.

The \emph{Scan} step computes asymmetric distances between the query vector and all codes stored in selected cells. Each ADC requires (1) $m$ accesses to cache-resident lookup tables and (2) $m$ additions. If implementing additions using SIMD is straightforward, SIMD does not allow an efficient implementation of table lookup, even using \texttt{gather} instructions introduced in recent processors~\cite{Andre2015,Hofmann2014}. SIMD can add 4 floating-point numbers (128 bits) or 8 floating-point numbers (256 bits) at once, there are only 2 cache read ports in each CPU core. Therefore, it is not possible to perform more than 2 cache accesses concurrently.

\begin{figure}
  \centering
  \begin{tikzpicture}
\node[register] (indexes)
{%
  \nodepart[elt]{one}$a_0$
  \nodepart[elt]{two}$b_0$
  \nodepart[elt]{three}$c_0$
  \nodepart[elt]{four}$\dots$%
  \nodepart[elt]{five}$p_0$%
};

\node[register, below=\vertSep of indexes.south, anchor=north] (table)
{%
  \nodepart[elt]{one}$D^0[0]$
  \nodepart[elt]{two}$D^0[1]$
  \nodepart[elt]{three}$D^0[2]$
  \nodepart[elt]{four}$\dots$%
  \nodepart[elt]{five}$D^0[15]$
};

\node[operator, below=\vertSep of table.south, anchor=north] (shuffle)
{\texttt{simd\_shuffle}};

\node[register, below=\vertSep of shuffle.south, anchor=north] (result)
{%
  \nodepart[elt]{one}$D^0[a_0]$
  \nodepart[elt]{two}$D^0[b_0]$
  \nodepart[elt]{three}$D^0[c_0]$
  \nodepart[elt]{four}$\dots$
  \nodepart[elt]{five}$D^0[p_0]$
};

\draw[->] (table.west) -- ++(-\horSep,0) |- (shuffle.west);
\draw[->] (indexes.east) -- ++(\horSep, 0) |- (shuffle.east);
\draw[->] (shuffle.south) -- (result.north);
\node[label, left=\labelSep of indexes.west, anchor=east] {$\mathit{indexes}$};
\node[label, left=\labelSep of table.west, anchor=east] {$\mathit{table}$};
\end{tikzpicture}
  \caption{SIMD in-register shuffle\label{fig:shuffle}}
\end{figure}
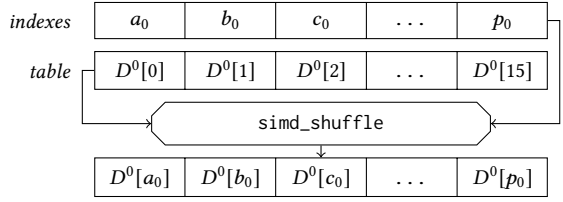

Therefore, efficiently implementing ADC using SIMD requires storing lookup tables in SIMD registers and performing lookups using SIMD in-register shuffles. The main challenge is that SIMD registers (128 bits) are much smaller than lookup tables, for common PQ configurations. In most cases, product quantizers use 8-bit sub-quantizers, which results in lookup tables of $k=2^{8}=256$ floats (8192 bits). For this reason, Quick ADC relies on (i) the use of 4-bit quantizers instead of the common 8-bit quantizers, and (ii) the quantization of floats to 8-bit integers. We obtain lookup tables of $k=2^{4}=16$ floats, which are then quantized to 8-bit integers. The resulting lookup tables comprise 16 8-bit integers (128 bits), and can be stored in SIMD registers. Once lookup tables are stored in SIMD registers, in-register shuffles can be used to perform 16 lookups in 1 cycle (Figure \ref{fig:shuffle}), enabling large performance gains.

In addition to the use of 4-bit quantizers and the quantization of floats to 8-bit integers, Quick ADC requires a minor change of memory layout. In the next sections, we detail this change of memory layout as well as our lookup tables quantization process and the SIMD implementation of distance computations.

\begin{figure}
  \centering
  \subfloat[Standard layout]{%
    \centering%
    \begin{tikzpicture}
\node [memory,
  column 1/.style={text=black!70},
] (layout)
{ a_1\,a_0 & \dots & a_{m-1}\,a_{m-2} \\
  b_1\,b_0 & \dots & b_{m-1}\,b_{m-2} \\
  \dots & \dots & \dots \\
  p_1\,p_0 & \dots & p_{m-1}\,p_{m-2} \\
};

  \node[label, left=\horSep of layout-1-1.west, anchor=east] {$a$};
  \node[label, left=\horSep of layout-2-1.west, anchor=east] {$b$};
  \node[label, left=\horSep of layout-4-1.west, anchor=east] {$p$};

\end{tikzpicture}%
    \label{fig:layoutnorm}%
  }\hfil%
  \subfloat[Transposed layout]{%
    \centering%
      \begin{tikzpicture}
  \node [memory, row 1/.style={text=black!70}] (layout)
  { a_1\,a_0 & b_1\,b_0 & c_1\,c_0 &\dots & p_1\,p_0 \\
    \dots & \dots & \dots & \dots & \dots  \\
    a_{m-1}\,a_{m-2} & b_{m-1}\,b_{m-2} & c_{m-1}\,c_{m-2} & \dots & a_{m-1}\,a_{m-2} \\
  };

  \node[label, above=\vertSep of layout-1-1.north, anchor=south] {$a$};
  \node[label, above=\vertSep of layout-1-2.north, anchor=south] {$b$};
  \node[label, above=\vertSep of layout-1-3.north, anchor=south] {$c$};
  \node[label, above=\vertSep of layout-1-5.north, anchor=south] {$p$};
  \end{tikzpicture}%
    \label{fig:layouttrans}%
  }%
  \caption{Inverted list memory layouts. Each table cell represents a byte.\label{fig:layout}}
\end{figure}

\subsection{Memory Layout}
\label{sec:memlayout}
An SIMD in-register shuffle performs 16 lookups at once, but in a \emph{single} lookup table \eg $D^0$ (Figure \ref{fig:shuffle}). Therefore, to use shuffles efficiently, we need to operate on the first component of 16 codes ($a_0,\dotsc,p_0$) at once instead of the 16 components of a single code ($a_{0},\dotsc,a_{15}$). Its is crucial for efficiency that all values in an SIMD register can be loaded in a single memory read. This requires that $a_0,\dotsc,p_0$ are \emph{contiguous} in memory, which is not the case with the standard memory layout of inverted lists (Figure \ref{fig:layoutnorm}). We therefore transpose inverted lists by blocks of 16 codes, so that analogous components of 16 codes are stored in adjacent bytes (Figure \ref{fig:layouttrans}). We divide each inverted list in blocks of 16 codes and transpose each block independently. Figure \ref{fig:layout} shows the transposition of one block of 16 codes ($a,\dotsc,p$).
This transposition is performed offline, and does not increase ANN query response time. The transposition is moreover very fast; the overhead on database creation time is less than 1\%.

\subsection{Quantization of Lookup Tables}
\label{sec:quantlook}

In standard ADC, lookup tables store 32-bit floats.  To be able to store tables of 16 elements in 128-bit registers, we quantize 32-bit floats to 8-bit integers using a scalar quantizer. Because there is no SIMD instruction to compare unsigned 8-bit integers, we quantize distances to signed 8-bit integers, only using their positive range. We quantize distances between a $\mathit{qmin}$ and $\mathit{qmax}$ bound into $n=127$ bins (0-126) uniformly. The size of each bin is $\Delta = (\mathit{qmax}-\mathit{qmin})/n$. Values larger than $\mathit{qmax}$ are quantized to 127.

We choose the minimum value accross all lookup tables $\{D^j\}_{j=0}^m$, which is the smallest distance we need to represent, as the $\mathit{qmin}$ value. Using the maximum possible distance \ie the sum of the maximums of all lookup tables results in a too high quantization error. Therefore, to set $\mathit{qmax}$ we scan $\mathit{init}$ vectors (typical $\mathit{init}$=200-1000) to find a temporary set of $R$ nearest neighbor candidates, where $R$ is the number of nearest neighbors requested by the user (Section \ref{sec:annsearch}). We use the distance of the query vector to the $R$th nearest neighbor candidate \ie the farthest nearest neighbor candidate, as the $\mathit{qmax}$ bound. All subsequent candidates will need to be closer to the query vector, thus $\mathit{qmax}$ is the maximum distance we need to represent.

\subsection{SIMD Distance Computation}

Although recent Intel CPUs offer 256-bit SIMD, we describe a version of Quick ADC which uses 128-bit SIMD for the sake of simplicity. Yet, we explain how to generalize it to 256-bit at the end of the section. Moreover, the 128-bit version of Quick ADC offers the best compatibility, notably with older Intel CPUs or ARM CPUs. In Algortihm~\ref{alg:quickadc}, SIMD instructions are denoted by the prefix \texttt{simd\_}. SIMD instructions use 128-bit variables, denoted by \texttt{r128}.

\begin{algorithm}[t]
  \caption{ANN Search with Quick ADC}\label{alg:quickadc}
  \begin{algorithmic}[1]
    \Function{lookup\_add}{$\mathit{comps}, D^j, \mathit{acc}$}  \Comment{Fig. \ref{fig:lookup_add}} \label{alg:line:lookadd}
    \State \texttt{r128} masked $\gets$ \texttt{simd\_and}($\mathit{comps}, \texttt{0x0f}$)
    \State \texttt{r128} partial $\gets$ \texttt{simd\_shuffle}($\mathit{comps}, D^j$)
    \State  \Return{\texttt{simd\_add\_saturated}($\mathit{acc}, \mathit{partial}$)}
    \EndFunction
    
    \Function{quick\_adc\_block}{$\mathit{blk}, \{D^j\}_{j=0}^{m-1}$}
    \State \texttt{r128} $\mathit{acc} \gets \{0\}$
    \For{$j \gets 0$ to $m/2 - 1$} \label{alg:line:rowiter}
    \State \texttt{r128} $\mathit{comps} \gets$ \texttt{simd\_load}($\mathit{blk} + j\cdot16$) \label{alg:line:rowload}
    \State $\mathit{acc} \gets$ \Call{lookup\_add}{$\mathit{comps}, D^{2j}, \mathit{acc}$} \label{alg:line:lookadd1}
    \State $\mathit{comps} \gets$ \texttt{simd\_right\_shift}($\mathit{comps}, 4$) \Comment{Fig. \ref{fig:shift}} \label{alg:line:shift}
    \State $\mathit{acc} \gets$ \Call{lookup\_add}{$\mathit{comps}, D^{2j+1}, \mathit{acc}$} \label{alg:line:lookadd2}
    \EndFor
    \Return{$\mathit{acc}$}
    \EndFunction
    
    \Function{quick\_adc\_scan}{$\mathit{tlist}, \{D^j\}_{j=0}^{m-1}, R$} \label{alg:line:qadc}
    \State $\mathit{neighbors} \gets \operatorname{binheap}(R)$
    \For{$\mathit{blk}$ in $\mathit{tlist}$} \label{alg:line:blockiter}
    \State \texttt{r128} $\mathit{acc} \gets$ \Call{quick\_adc\_block}{$\mathit{blk}, \{D^j\}_{j=0}^{m-1}$}
    \State \Call{extract\_matches}{$\mathit{acc}, \mathit{neighbors}$} \label{alg:line:extract}
    \EndFor
    \State \Return{$\mathit{neighbors}$}
    \EndFunction
  \end{algorithmic}
\end{algorithm}

The \textsc{quick\_adc\_scan} function (Algorithm~\ref{alg:quickadc}, line~\ref{alg:line:qadc}) scans a block-transposed inverted list $\mathit{tlist}$ (Section~\ref{sec:memlayout}) using $m$ \emph{quantized} lookup tables $\{D^j\}_{j=0}^{m-1}$, where $m$ is the number of sub-quantizers of the product quantizer. Each lookup table is stored in a distinct SIMD register. The \textsc{quick\_adc\_scan} function iterates over blocks $\mathit{blk}$ of 16 codes (Algorithm~\ref{alg:quickadc}, line~\ref{alg:line:blockiter}). The \textsc{quick\_adc\_block} function computes the distance between the query vector and the 16 codes ($a,\dotsc,p$) of the block $\mathit{blk}$.

\begin{figure}
  \centering
  \begin{tikzpicture}
\node[register] (indexes01)
{%
  \nodepart[elt]{one}$a_1\,a_0$
  \nodepart[elt]{two}$b_1\,b_0$
  \nodepart[elt]{three}$c_1\,c_0$
  \nodepart[elt]{four}$\dots$%
  \nodepart[elt]{five}$p_1\,p_0$%
};

\node[operator, below=\vertSep of indexes01.south, anchor=north] (shift) {\texttt{simd\_right\_shift} (4 bits)};

\draw[->] (indexes01.west) -- ++(-\horSep,0) |- (shift.west);

\node[register, below=\vertSep of shift.south] (shiftindexes01)
{%
  \nodepart[elt]{one}$a_1$
  \nodepart[elt]{two}$a_0\,b_1$
  \nodepart[elt]{three}$b_0\,c_1$
  \nodepart[elt]{four}$\dots$%
  \nodepart[elt]{five}$o_0\,p_1$%
};

\draw[->] (shift.south) -- (shiftindexes01.north);

\node[label, left=\labelSep of shiftindexes01.west, anchor=east] {$\mathit{comps}$};
\node[label, left=\labelSep of indexes01.west, anchor=east] {$\mathit{comps}$};

\node[label, anchor=west, right=1cm of shift.east] (prev) {\phantom{$\mathit{acc}$}};
\node[label, left=\labelSep of indexes01.west, anchor=east] {\phantom{$\mathit{masked}$}};

\end{tikzpicture}
  \caption{SIMD 4-bit Right Shift\label{fig:shift} ($j=0$)}
\end{figure}
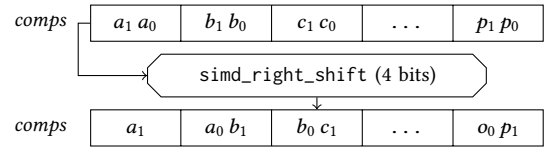

Each block comprises $m/2$ rows of 16 bytes (128 bits). Each row stores the $j$th and $(j+1)$th components of 16 codes (Figure~\ref{fig:layouttrans}). The \textsc{quick\_adc\_block} function iterates over each row (Alorithm \ref{alg:quickadc}, line \ref{alg:line:rowiter}), and loads it in the $\mathit{comps}$ register sequentially (Algorithm~\ref{alg:quickadc}, line \ref{alg:line:rowload}). Two lookup-add operations are performed on each row (Algorithm \ref{alg:quickadc}, line \ref{alg:line:lookadd1} and line~\ref{alg:line:lookadd2}): one for the $(2j)$th components, and one for $(2j+1)$th components of the codes. Figure \ref{fig:lookup_add} describes the succession of operations performed by the \textsc{lookup\_add} function for the first row ($j=0$). As each byte of the first row stores two components, \eg the first byte of the first row stores $a_1$ and $a_0$ (Figure~\ref{fig:lookup_add}), we start by masking the lower 4 bits of each byte (\texttt{and} with \texttt{0x0f}), to obtain the first components ($a_0,\dotsc,p_0$) only. The remainder of the function looks up values in the $D^0$ table and accumulates distances in $\mathit{acc}$ variable. Before the \textsc{lookup\_add} function can be used to process the second components ($a_1,\dotsc,p_1$), it is necessary that ($a_1,\dotsc,p_1$) are in the lowest 4 bits of each byte of the register. We therefore right shift the $\mathit{comps}$ register by 4 bits (Figure~\ref{fig:shift}) before calling \textsc{lookup\_add} (Algorithm~\ref{alg:quickadc}, line~\ref{alg:line:shift}). The \textsc{extract\_matches} function (Algorithm~\ref{alg:quickadc}, line~\ref{alg:line:extract}), the implementation of which is not shown, extracts distances from the $\mathit{acc}$ register and inserts them in the binary heap $\mathit{neighbors}$. 

Among 256-bit SIMD instructions (AVX and AVX2 instruction sets) supported on recent CPUs, some, like in-register shuffles, operate concurrently on two independent 128-bit lanes. This prevents use of 256-bit lookup tables (32 8-bit integers) but allows an easy generalization of the 128-bit version of Quick ADC. While the 128-bit version of Quick ADC iterates on block rows one by one (Algorithm~\ref{alg:quickadc}, line~\ref{alg:line:rowiter}), the 256-bit version processes two rows at once: one row in each 128-bit lane. The number of iterations is thus reduced from $m/2$ to $m/4$. Lastly, instead of storing each $D^j$ table in a distinct 128-bit register, the tables $D^j$ and $D^{2j}$, $j \in \{0,\dotsc,m/2-1\}$, are stored in each of the two lanes of a 256-bit register.

\begin{figure}
  \centering
  \begin{tikzpicture}

\node[register] (indexes01)
{%
  \nodepart[elt]{one}$a_1\,a_0$
  \nodepart[elt]{two}$b_1\,b_0$
  \nodepart[elt]{three}$c_1\,c_0$
  \nodepart[elt]{four}$\dots$%
  \nodepart[elt]{five}$p_1\,p_0$%
};

\node[operator, below=\vertSep of indexes01.south, anchor=north] (mask0) {\texttt{simd\_and} (\texttt{0x0f})};

\draw[->] (indexes01.west) -- ++(-\horSep,0) |- (mask0.west);

\node[register, below=\vertSep of mask0.south, anchor=north] (indexes0)
{%
  \nodepart[elt]{one}$a_0$
  \nodepart[elt]{two}$b_0$
  \nodepart[elt]{three}$c_0$
  \nodepart[elt]{four}$\dots$%
  \nodepart[elt]{five}$p_0$%
};

\draw[->] (mask0.south) -- (indexes0.north);

\node[label, left=\labelSep of indexes01.west, anchor=east] {$\mathit{comps}$};
\node[label, left=\labelSep of indexes0.west, anchor=east] {$\mathit{masked}$};
\node[register, below=\vertSep of indexes0.south, anchor=north] (tables0)
{%
  \nodepart[elt]{one}$D^0[0]$
  \nodepart[elt]{two}$D^0[1]$
  \nodepart[elt]{three}$D^0[2]$
  \nodepart[elt]{four}$\dots$%
  \nodepart[elt]{five}$D^0[15]$%
};

\node[operator, below=\vertSep of tables0.south, anchor=north] (shuffle0) {\texttt{simd\_shuffle}};

\draw[->] (indexes0.west) -- ++(-\horSep,0) |- (shuffle0.west);
\draw[->] (tables0.east) -- ++(\horSep,0) |- (shuffle0.east);

\node[register, below=\vertSep of shuffle0.south, anchor=north] (look0)
{%
  \nodepart[elt]{one}$D^0[a_0]$
  \nodepart[elt]{two}$D^0[b_0]$
  \nodepart[elt]{three}$D^0[c_0]$
  \nodepart[elt]{four}$\dots$%
  \nodepart[elt]{five}$D^0[p_0]$%
};

\draw[->] (shuffle0.south) -- (look0.north);
\node[label, left=\labelSep of look0.west, anchor=east] {$\mathit{partial}$};

\node[operator, below=\vertSep of look0.south, anchor=north] (add0) {
  \texttt{simd\_add\_saturated}};

\draw[->] (look0.west) -- ++(-\horSep,0) |- (add0.west);

\node[register, below=\vertSep of add0.south, anchor=north] (res0)
{%
  \nodepart[elt]{one}$\mathit{acc[0]} + D^0[a_0]$
  \nodepart[elt]{two}$\mathit{acc[1]} + D^0[b_0]$
  \nodepart[elt]{three}$\mathit{acc[2]} + D^0[c_0]$
  \nodepart[elt]{four}$\dots$%
  \nodepart[elt]{five}$\mathit{acc[15]} + D^0[p_0]$%
};

\node[label, left=\labelSep of res0.west, anchor=east] {$\mathit{acc}$};

\draw[->] (add0.south) -- (res0.north);
\node[label, anchor=west, right=1cm of add0.east] (prev) {$\mathit{acc}$};
\draw [->] (prev.west) -- (add0.east);

\end{tikzpicture}
  \caption{SIMD Lookup-add\label{fig:lookup_add} ($j=0$)}
\end{figure}
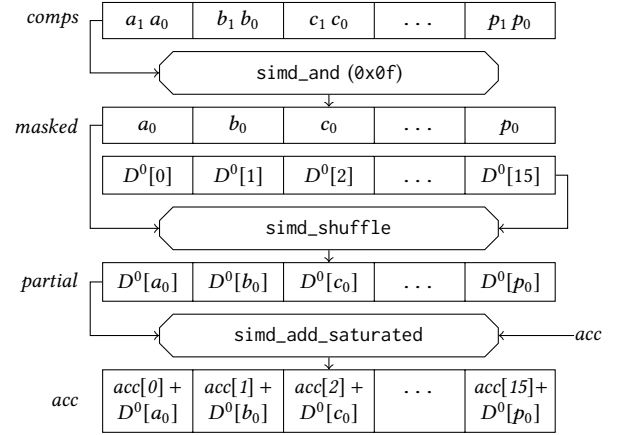

\section{Evaluation}
\label{sec:eval}

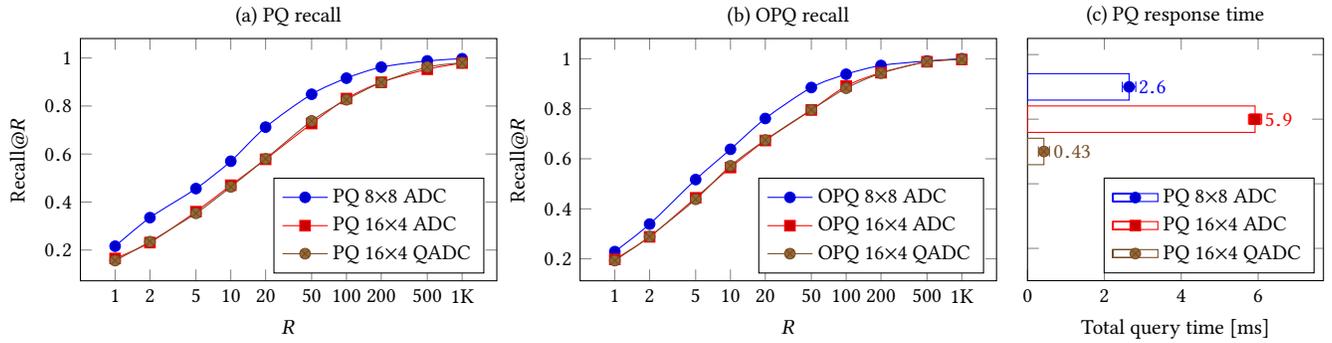
\begin{figure*}
  \centering
   \begin{tikzpicture}
   \begin{semilogxaxis}[
   name=recall pq,
   title=(a) PQ recall,
   recall plot,
   width=0.4*\linewidth,
   height=0.27*\linewidth,
   ]
   \addplot+[smooth] 
   table[x=r,y=pq8x8adc] {graphs/SIFT1M.flat.recall.gnuplot};
   \addplot+[smooth] 
   table[x=r,y=pq16x4adc] {graphs/SIFT1M.flat.recall.gnuplot};
   \addplot+[smooth] 
   table[x=r,y=pq16x4qadc] {graphs/SIFT1M.flat.recall.gnuplot};
   \legend{PQ $8{\stimes}8$ ADC, PQ $16{\stimes}4$ ADC, PQ $16{\stimes}4$ QADC}
   \end{semilogxaxis}
   \begin{semilogxaxis}[
   name=recall opq,
   title=(b) OPQ recall,
   recall plot,
   width=0.4*\linewidth,
   height=0.27*\linewidth,
   at={($(recall pq.east)+(3.5em,0cm)$)},
   anchor=west,
   ]
   \addplot+[smooth] 
   table[x=r,y=opq8x8adc] {graphs/SIFT1M.flat.recall.gnuplot};
   \addplot+[smooth] 
   table[x=r,y=opq16x4adc] {graphs/SIFT1M.flat.recall.gnuplot};
   \addplot+[smooth] 
   table[x=r,y=opq16x4qadc] {graphs/SIFT1M.flat.recall.gnuplot};
   \legend{OPQ $8{\stimes}8$ ADC, OPQ $16{\stimes}4$ ADC, OPQ $16{\stimes}4$ QADC}
   \end{semilogxaxis}
   \begin{axis}[
   time plot,
   title=(c) PQ response time,
   at={($(recall opq.east)+(1.3em,0cm)$)},
   anchor=west,
   width=0.31*\linewidth,
   height=0.27*\linewidth,
   ymin = -5,
   ymax = 2.5,
   xmin = 0,
   xmax=7.7
   ]
   \addplot+[time bar]
   table[x=pq8x8adcmean,
   x error=pq8x8adcstd,
   y expr=1,
   meta=pq8x8adcmean]
   {graphs/SIFT1M.flat.time.transposed.gnuplot};
   \addplot+[time bar]
   table[x=pq16x4adcmean,
   x error=pq16x4adcstd,
   y expr=0,
   meta=pq16x4adcmean]
   {graphs/SIFT1M.flat.time.transposed.gnuplot};
   \addplot+[time bar]
   table[x=pq16x4qadcmean,
   x error=pq16x4qadcstd,
   y expr=-1,
   meta=pq16x4qadcmean]
   {graphs/SIFT1M.flat.time.transposed.gnuplot};
   \legend{PQ $8{\stimes}8$ ADC, PQ $16{\stimes}4$ ADC, PQ $16{\stimes}4$ QADC}
   \end{axis}
   \end{tikzpicture}
  \caption{ADC and QADC response time and recall with PQ and OPQ (SIFT1M, Exhaustive search)\label{fig:exhaust}}
\end{figure*}

\subsection{Experimental Setup}

We implemented 256-bit Quick ADC in C++, using compiler intrinsics to access SIMD instructions. Our implementation is released under the Clear BSD license\footnote{\url{https://github.com/technicolor-research/quick-adc}} and uses the AVX and AVX2 instruction sets. We used the g++ compiler version 5.3, with the options \texttt{-03 -ffast-math -m64 -march=native}. Exhaustive search and non-exhaustive search (inverted indexes, IVF) were implemented as described in~\cite{Jegou2011}. We use the yael library and the ATLAS library version 3.10.2. We compiled an optimized version of ATLAS on our system. To learn product quantizers and optimized product quantizers, we used the implementation \footnote{\url{https://github.com/arbabenko/Quantizations}} of the authors of~\cite{Babenko2015TQ, Babenko2014AQ}.
Unless otherwise noted, experiments were performed on our workstation (Table~\ref{tbl:platforms}). To get accurate timings, we processed queries sequentially on a single core. We evaluate our approach on two publicly available\footnote{\url{http://corpus-texmex.irisa.fr/}} datasets of SIFT descriptors, one dataset of GIST descriptors, and one dataset of PCA-compressed deep features\footnote{\url{http://sites.skoltech.ru/compvision/projects/aqtq/}} (Table~\ref{tbl:datasets}). For SIFT1B, the learning set is needlessly large to train product quantizers, so we used the first 2 million vectors. We used a query set of 1000 vectors for all experiments.

\begin{table}
  \centering
  \caption{Systems\label{tbl:platforms}}
  \begin{tabularx}{\linewidth}{Xll@{  }l}
    \toprule
    & CPU &  \multicolumn{2}{l}{RAM} \\
    \midrule
    workstation & Xeon E5-1650v3 & 16GB & DDR4 2133Mhz\\
    server & Xeon E5-2630v3 & 128GB & DDR4 1866Mhz\\
    \bottomrule
  \end{tabularx}
\end{table}

\begin{table}
  \centering
  \caption{Datasets\label{tbl:datasets}}
  \begin{tabularx}{\linewidth}{lXXXl}
    \toprule
    & Base set & Learning set & Query set & Dim. \\
    \midrule
    SIFT1M & 1M & 100K & 10K (1K) & 128\\
    SIFT1B & 1000M & 100M (2M) & 10K (1K) & 128 \\
    GIST1M & 1M & 500K & 1K & 960 \\
    Deep1M & 1M & 300K & 1K & 256\\
    \bottomrule
  \end{tabularx}
\end{table}

\subsection{Exhaustive Search in SIFT1M}

Using $16{\stimes}4$ Quick ADC (QADC) instead of $8{\stimes}8$ ADC offers a large performance gain, thanks to the use of SIMD in-register shuffles. It however also causes a decrease in recall which is cause by two factors: (1) use of $16{\stimes}4$ quantizers instead of $8{\stimes}8$ quantizers (Section~\ref{sec:paramimpact}) and (2) use of quantized lookup tables (Section~\ref{sec:quantlook}). In this section, we evaluate the global decrease in recall caused by the use of $16{\stimes}4$ QADC instead of $8{\stimes}8$ ADC, but also the relative impact of factors (1) and (2). To do so, we use the SIFT1M dataset and follow an exhaustive search strategy. We do not use an inverted index and we encode the original vectors into short codes, not residuals. This maximizes quantization error and thus represents a worst-case scenario for QADC. We scan $\mathit{init}=200$ vectors to set the $\mathit{qmax}$ bound for quantization of lookup tables (Section ~\ref{sec:quantlook}).

We observe that $16{\stimes}4$ ADC slightly decreases recall (Figure~\ref{fig:exhaust}a). However, $16{\stimes}4$ QADC, which uses quantized lookup tables, does not further decrease recall in comparison with $16{\stimes}4$ ADC. OPQ yields better results than PQ in all cases (Figure~\ref{fig:exhaust}b), which is consistent with~\cite{Norouzi2013, Ge2014}. Moreover, the difference in recall between $8{\stimes}8$ ADC and $16{\stimes}4$ QADC is lower for OPQ than it is for PQ. OPQ optimizes the decomposition of the input vector space into $m$ sub-spaces, which are used by the optimized product quantizer (Section \ref{sec:vecenc}). For $m=16$, OPQ has more degrees of freedom than for $m=8$ and is therefore able to bring a greater level of optimization.

For an exhaustive search in 1 million vectors, $16{\stimes}4$ QADC is ${\sim}14$ times faster than $16{\stimes}4$ ADC and $6$ times faster than $8{\stimes}8$ ADC (Figure~\ref{fig:exhaust}c) (85\% decrease in response time). Response times for PQ and OPQ are similar, so we report results for PQ.
In practice, $8{\stimes}8$ ADC is much more common than $16{\stimes}4$ ADC~\cite{Babenko2014AQ, Babenko2015TQ, Babenko2015, Norouzi2013, Zhang2014}, thus we only compare $16{\stimes}4$ QADC with $8{\stimes}8$ ADC in the remainder of this section. Overall, QADC therefore proposes trading a small decrease in recall, for a large improvement in response time.

Non-exhaustive search offers both a lower response time and a higher recall than exhaustive search (Section~\ref{sec:ivf}). For this reason, non-exhaustive search is preferred to exhaustive search in practical systems. Therefore, in the remainder of this section, we evaluate QADC in the context of non-exhaustive search, for a wide range of scenarios: SIFT, GIST descriptors, deep feature, PQ and OPQ, 64 and 128 bit codes. We show that in most cases, when combined with OPQ and inverted indexes, QADC offers a decrease in response time close to 70\% for a small or negligible loss of accuracy.

\subsection{Non-exhaustive Search in SIFT1M}
\label{sec:sift1m}

Table~\ref{tbl:sift1m} compares the Recall@100 (R@100) and total ANN search time (Total). The time spent in each of the search steps (Index, Tables, and Scan) detailed in Section~\ref{sec:annsearch} is also reported. All times are in milliseconds (ms). OPQ requires a rotation of the input vector before computing lookup tables (Section~\ref{sec:vecenc}). We include the time to perform this rotation in the Tables column. When using inverted indexes, the parameters $K$, the total number of cells of the inverted index, and $\mathit{ma}$, the number of cells scanned to answer a query, impact response time and recall (Section~\ref{sec:annsearch}). For datasets of 1 million vectors, we have found the parameters $\mathit{ma}=24$ and $K=256$ to offer the best tradeoff.

For this configuration, QADC offers a 75\% decrease in scan time. In addition, QADC offers a 50-70\% decrease in tables computation time, thanks to the use of 4-bit quantizers, which result in smaller and faster to compute small tables. Overall, this translates into a decrease of approximately 70\% in total response time. The loss of recall is significantly lower with OPQ (-1.5\%) than with PQ (-4.4\%), as OPQ offers a lower quantization error than PQ.

\makeatletter
\newcommand\primitiveinput[1]
{\input{#1} }
\makeatother

\begin{table}
\begin{threeparttable}
\caption{Non-exhaustive search, SIFT1M, 64 bit\label{tbl:sift1m}}
\begin{tabularx}{\linewidth}{lXlllll}
  \toprule
  PQ & ADC \tnote{*} & R@100 & Index & Tables & Scan & Total \\
  \midrule
  %
  %
  %
  \multicolumn{7}{c}{\textbf{SIFT1M, IVF, K=256, ma=24}} \\
  \midrule
  PQ & ADC & 0.949 & 0.008 & 0.18 & 0.3 & 0.48 \\ 
    & QADC & 0.907 & 0.008 & 0.055 & 0.072 & 0.14 \\ 
    &      & \emph{-4.4\%} &    & \emph{-69\%} & \emph{-76\%} & \emph{-72\%} \\ 
\midrule
OPQ & ADC & 0.963 & 0.008 & 0.21 & 0.29 & 0.52 \\ 
    & QADC & 0.949 & 0.008 & 0.089 & 0.073 & 0.17 \\ 
    &      & \textbf{\emph{-1.5\%}} &    & \emph{-59\%} & \emph{-75\%} & \textbf{\emph{-67\%}} \\ 

  \bottomrule
\end{tabularx}
\begin{tablenotes}
  \small
  \item[*] ADC: $8{\stimes}8$ ADC, QADC: $16{\stimes}4$ QADC
\end{tablenotes}
\end{threeparttable}
\end{table}

\subsection{Non-exhaustive Search in GIST1M}
\label{sec:gist1m}

GIST descriptors have a much higher dimensionality (960 dimensions) than SIFT descriptors (128 dimensions). For this reason, GIST descriptors are often encoded into 128-bit codes instead of 64-bit codes~\cite{Norouzi2013, Babenko2015}. This corresponds to $16{\stimes}8$ codes for ADC and $32{\stimes}4$ codes for QADC. In this case, when combined with PQ, QADC offers a decrease of approximately 70\% in total response time, as with 64-bit codes and SIFT descriptors (Table~\ref{tbl:gist1m128}). However, the decrease in recall when using QADC with PQ is higher for GIST descriptors (-24\%) than for SIFT descriptors (-4.4\%). This loss of recall is limited to 5\% when combining QADC with OPQ. The decrease in total response time is however less important for OPQ (-45\%) than for PQ (-71\%). This is because OPQ requires rotating the query vector when computing distance tables. Rotating 960-dimensional GIST descriptors is costly, increasing the time to compute distance tables, which in turn limits the gain in total response time.

\begin{table}
  \begin{threeparttable}
    \caption{Non-exhaustive search, GIST1M, 128 bit\label{tbl:gist1m128}}
    \begin{tabularx}{\linewidth}{lXlllll}
      \toprule
      PQ & ADC \tnote{*} & R@100 & Index & Tables & Scan & Total \\
      \midrule
      \multicolumn{7}{c}{\textbf{GIST1M, IVF, K=256, ma=24}} \\
      \midrule
      PQ & ADC & 0.675 & 0.038 & 0.77 & 0.71 & 1.5 \\ 
    & QADC & 0.515 & 0.038 & 0.26 & 0.15 & 0.45 \\ 
    &      & \emph{-24\%} &    & \emph{-67\%} & \emph{-79\%} & \emph{-71\%} \\ 
\midrule
OPQ & ADC & 0.918 & 0.039 & 1.7 & 0.73 & 2.5 \\ 
    & QADC & 0.872 & 0.038 & 1.2 & 0.16 & 1.4 \\ 
    &      & \textbf{\emph{-5\%}} &    & \emph{-32\%} & \emph{-78\%} & \textbf{\emph{-45\%}} \\ 

      \bottomrule
    \end{tabularx}
    \begin{tablenotes}
      \small
      \item[*] ADC: $16{\stimes}8$ ADC, QADC: $32{\stimes}4$ QADC
    \end{tablenotes}
  \end{threeparttable}
\end{table}

\subsection{Non-exhaustive Search in Deep1M}
\label{sec:deep1m}

The Deep1M dataset comprises $L_2$-normalized deep features that are PCA-compressed to 256 dimensions. Due to their relatively low dimensionality, these vectors can be encoded into 64-bit codes. As for SIFT and GIST descriptors, QADC offers a 70\% decrease in response time when combined with PQ. The decrease in recall (-13\%) for Deep1M vectors is between the decrease in recall for SIFT descriptors (-4.4\%) and the decrease in recall for GIST descriptors (-24\%), which is consistent with the dimensionality of vectors. Once again, OPQ strongly limits the loss of recall (-2.2\%). It also limits the gain in total response time (-62\%), due to the time spent performing rotations.

\begin{table}
  \begin{threeparttable}
    \caption{Non-exhaustive search, Deep1M, 64 bit\label{tbl:deep1m}}
    \begin{tabularx}{\linewidth}{lXlllll}
      \toprule
      PQ & ADC \tnote{*} & R@100 & Index & Tables & Scan & Total \\
      \midrule
      \multicolumn{7}{c}{\textbf{Deep1M, IVF, K=256, ma=24}} \\
      \midrule
      PQ & ADC & 0.772 & 0.015 & 0.24 & 0.33 & 0.58 \\ 
    & QADC & 0.669 & 0.013 & 0.082 & 0.076 & 0.17 \\ 
    &      & \emph{-13\%} &    & \emph{-66\%} & \emph{-77\%} & \emph{-71\%} \\ 
\midrule
OPQ & ADC & 0.922 & 0.013 & 0.34 & 0.32 & 0.67 \\ 
    & QADC & 0.902 & 0.013 & 0.16 & 0.08 & 0.25 \\ 
    &      & \emph{-2.2\%} &    & \emph{-53\%} & \emph{-75\%} & \emph{-62\%} \\ 

      \bottomrule
    \end{tabularx}
    \begin{tablenotes}
      \small
      \item[*] ADC: $8{\stimes}8$ ADC, QADC: $16{\stimes}4$ QADC
    \end{tablenotes}
  \end{threeparttable}
\end{table}

\subsection{Non-exhaustive Search in SIFT1B}
\label{sec:sift1b}

We conclude our experimental section by performing experiments on a large dataset of 1 billion SIFT descriptors. For this dataset, we use an inverted index with $K=65536$ cells and scan $\mathit{ma}=64$ cells to answer queries. This configuration has been shown to offer best performance~\cite{Babenko2015}. We scan $\mathit{init}=1000$ vectors before quantizing lookup tables (Section~\ref{sec:quantlook}). For SIFT descriptors, we have shown that combining QADC with OPQ allows a higher recall, with no impact on total response time (Section~\ref{sec:sift1m}). For this reason, we perform experiments with OPQ. As with the SIFT1M dataset, QADC offers a decrease of approximately 70\% in response time (Table~\ref{tbl:sift1b}). However, for 64-bit codes the loss of recall is higher on SIFT1B (-7.3\%) than on SIFT1M (-1.5\%). Using 128-bit codes makes the loss of recall negligible (-1.1\%) but increases memory use (Table~\ref{tbl:sift1b128}). With 128-bit codes, the database uses $20$GB of RAM, therefore we had to run this experiment on our server (Table~\ref{tbl:platforms}).

\subsection{Summary}

Our experiments show that QADC offers large performance gains in all cases, at the expense of a small or negligible decrease in recall. For SIFT descriptors, this decrease is always negligible, even when using PQ (Section~\ref{sec:sift1m}). For descriptors of higher dimensionality, the decrease in recall is greater  (Section~\ref{sec:gist1m} and~\ref{sec:deep1m}). Using OPQ brings this loss of recall down to low levels, but also slightly limits the gain in response time (-50\% to -60\% decrease in response time). Overall, QADC offers an interesting speed-accuracy tradeoff: a loss in recall of 1-5\% for a decrease of response time of 50-70\%. The case of large datasets is slightly less favorable for QADC: we observe a 7\% decrease in accuracy on the SIFT1B dataset, even when using OPQ. However, even in this scenario, QADC exhibits similar or better performance than state-of-the-art systems. Thus, with 64-bit codes, QADC achieves a recall of 0.747 on the SIFT1B dataset in 1.7 ms, while the state-of-the-art OMulti-D-OADC system achieves the same recall in 2 ms~\cite{Babenko2015}. With 128-bit codes, QADC achieves a recall of 0.94 in 3.4 ms while OMulti-D-OADC achieves a recall of 0.901 in 5 ms~\cite{Babenko2015}.

\begin{table}
  \begin{threeparttable}
    \caption{Non-exhaustive search, SIFT1B, 64 bit\label{tbl:sift1b}}
    \begin{tabularx}{\linewidth}{lXlllll}
      \toprule
      PQ & ADC \tnote{*} & R@100 & Index & Tables & Scan & Total \\
      \midrule
      \multicolumn{7}{c}{\textbf{SIFT1B, IVF, K=65536, ma=64}} \\
      \midrule
      OPQ & ADC & 0.806 & 0.52 & 0.51 & 4.2 & 5.2 \\ 
    & QADC & 0.747 & 0.53 & 0.22 & 0.92 & 1.7 \\ 
    &      & \textbf{\emph{-7.3\%}} &    & \emph{-57\%} & \emph{-78\%} & \textbf{\emph{-68\%}} \\ 

      \bottomrule
    \end{tabularx}
    \begin{tablenotes}
      \small
      \item[*] ADC: $8{\stimes}8$ ADC, QADC: $16{\stimes}4$ QADC
    \end{tablenotes}
  \end{threeparttable}
\end{table}

\begin{table}
  \begin{threeparttable}
    \caption{Non-exhaustive search, SIFT1B, 128 bits\label{tbl:sift1b128}}
    \begin{tabularx}{\linewidth}{lXlllll}
      \toprule
      PQ & ADC \tnote{*} & R@100 & Index & Tables & Scan & Total \\
      \midrule
      \multicolumn{7}{c}{\textbf{SIFT1B, IVF, K=65536, ma=64}} \\
      \midrule
      OPQ & ADC & 0.95 & 0.73 & 1.1 & 10 & 12 \\ 
    & QADC & 0.94 & 0.72 & 0.49 & 2.2 & 3.4 \\ 
    &      & \textbf{\emph{-1.1\%}} &    & \emph{-57\%} & \emph{-78\%} & \textbf{\emph{-72\%}} \\ 

      \bottomrule
    \end{tabularx}
    \begin{tablenotes}
      \small
      \item[*] ADC: $16{\stimes}8$ ADC, QADC: $32{\stimes}4$ QADC
    \end{tablenotes}
  \end{threeparttable}
\end{table}

\section{Related Work}
\emph{PQ Fast Scan.} Both PQ Fast Scan~\cite{Andre2015} and Quick ADC speed up the scan of lists of short codes by taking advantage of SIMD. More specifically, PQ Fast Scan and Quick ADC store lookup tables in SIMD registers and use SIMD in-register shuffles in place of cache accesses. The main challenge is that lookup tables used for ANN search are much larger than SIMD registers. PQ Fast Scan tackles this issue by both altering (i) inverted lists (vector grouping) and (ii) lookup tables (minimum tables). PQ Fast Scan uses the standard 8-bit sub-quantizers, hence it incurs no loss of recall. However, because of the transformations it applies, PQ Fast Scan requires a minimum list size of 3 million codes~\cite{Andre2015}. This limits the applicability of PQ Fast Scan to exhaustive search or to very coarse (and thus inefficient~\cite{Babenko2015, Jegou2011}) inverted indexes. On the contrary, Quick ADC imposes no constraints on list sizes, and may be combined with efficient inverted indexes (lists of 1000-10000 vectors) as shown in our experiments (Section~\ref{sec:eval}).

\emph{Inverted Multi-index.} Inverted multi-indexes~\cite{Babenko2015} provide a finer partition (typical $K=2^{28}$) of the vector space than inverted indexes (typical $K=65536$). At query time, this finer partition allows scanning less vectors to achieve the same recall, therefore providing a significant speedup. Unlike PQ Fast Scan, Quick ADC does not require lists of a minimum size. Therefore, Quick ADC can also be combined with multi-indexes to further decrease response time.

\emph{Compositional Quantization Models.} Recently, compositional vector quantization models inspired by PQ have been proposed. These models offer a lower quantization error than PQ or OPQ. Among these models are Additive Quantization (AQ) \cite{Babenko2014AQ}, Tree Quantization (TQ) \cite{Babenko2015TQ} and Composite Quantization (CQ) \cite{Zhang2014}. These models also use cache-resident lookup tables to compute distances, therefore Quick ADC may be combined with them. However, this may require additional work as some of these models use more lookup tables than PQ or OPQ.

\section{Conclusion}
In this paper, we presented Quick ADC, a novel distance computation method for ANN search. Quick ADC achieves a 3 to 6 times speedup over the standard ADC method by efficiently exploiting SIMD. This efficient use of SIMD is enabled by two changes to the ADC procedure: (i) the use of 4-bit quantizers instead of the usual 8-bit quantizers, and (ii) the quantization of floating-point distances to 8-bit integers.

It is known that using 4-bit quantizers may cause a loss in recall~\cite{Jegou2011}. However, through an extensive evaluation, we have shown that this loss is small or negligible when 4-bit quantizers are combined with OPQ and inverted indexes. In addition, we have shown that Quick ADC integrates well with other search acceleration methodes, in particular inverted indexes. Lastly, upcoming SIMD instruction sets (\eg 512-bit AVX in Xeon Skylake CPUs), will allow Quick ADC to offer both greater speedups (twice more codes processed per cycle) and an even smaller loss of recall (6-bit quantizers instead of 4-bit quantizers).

\section*{Acknowledgements}
Experiments presented in this paper were carried out using the Grid'5000 testbed, supported by a scientific interest group hosted by Inria and including CNRS, RENATER and several Universities as well as other organizations (see \url{https://www.grid5000.fr}). 

\bibliographystyle{ACM-Reference-Format}
\bibliography{references}

\end{document}